\documentclass[twoside,11pt]{article}

\usepackage{jmlr2e}
\usepackage[]{algorithm2e}
\usepackage{subfig}
\usepackage{amsmath}

\ShortHeadings{Gaussian Process Policy Optimization}{Rao, Sarkar, and Narayanan}
\firstpageno{1}

\begin{document}

\title{Gaussian Process Policy Optimization}

\author{\name Ashish Rao \email ashish.arartc@gmail.com \\
       \addr Cupertino High School\\
       Cupertino, CA 95014, USA
       \AND
       \name Bidipta Sarkar \email bidiptas13@gmail.com \\
       \addr Cupertino High School\\
       Cupertino, CA 95014, USA
       \AND 
       \name Tejas Narayanan \email tejasn100@gmail.com \\
       \addr Cupertino High School\\
       Cupertino, CA 95014, USA}

\editor{}

\maketitle

\begin{abstract}

We propose a novel actor-critic, model-free reinforcement learning algorithm which employs a Bayesian method of parameter space exploration to solve environments. A Gaussian process is used to learn the expected return of a policy given the policy's parameters. The system is trained by updating the parameters using gradient descent on a new surrogate loss function consisting of the Proximal Policy Optimization ‘Clipped’ loss function and a bonus term representing the expected improvement acquisition function given by the Gaussian process. This new method is shown to be comparable to and at times empirically outperform current algorithms on environments that simulate robotic locomotion using the MuJoCo physics engine.
\end{abstract}


\section{Introduction}

Reinforcement learning involves the study of how intelligent systems (or agents) can be made to learn behaviors (or policies) that yield high rewards in a given environment. Popular algorithms to tackle reinforcement learning problems include Policy Gradient methods like Proximal Policy Optimization (PPO) ~\citep{DBLP:journals/corr/SchulmanWDRK17} and Trust Region Policy Optimization (TRPO) ~\citep{DBLP:journals/corr/SchulmanLMJA15}, as well as techniques such as Deep Q Learning ~\citep{mnih2013playing}. Often times, deep neural networks are employed to represent certain quantities specific to each type of algorithm: the policy for Policy Gradient Methods, the policy and value function for Actor-Critic Methods (specific type of Policy Gradient methods), and the Q values for Q learning.

Despite various successes of these methods on a wide variety of environments, they possess certain shortcomings as described in ~\citep{DBLP:journals/corr/SchulmanWDRK17}. Deep Q Learning is constrained to problems involving discrete and low dimensional action spaces, can fail on simple problems, and is poorly understood. TRPO and some other policy gradient methods are complicated to implement and incompatible with certain model architectures. Simpler policy gradient methods, such as the vanilla policy gradient, are not data efficient. 

Besides the learning algorithm itself, it is vital for an agent to strike an effective balance between exploring its environment and gathering further information as to which actions in which states yield high rewards, and exploiting that knowledge to achieve those high rewards. Deep Reinforcement Learning methods often use entropy regularization to achieve this balance, but this technique can be ineffective in higher dimensional action spaces involving sparse rewards ~\citep{nachum2016improving}. In addition, recent work has shown that rather than noise injection in the action space, adding noise noise in the parameter space can yield richer behaviors ~\citep{plappert2017parameter}. 

We propose a new algorithm which we call GPPO, or Gaussian Process Policy Optimization. It utilizes a Gaussian Process, a Bayesian Model, to encourage exploration in the parameter space of parameters that maximize a given acquisition function, a weighted sum of the predicted return of a policy with those parameters and the variance of the Gaussian Process. Adding this acquisition function as a bonus to the PPO Clipped loss provides at times improved performance over current state-of-the-art methods.

\section{Preliminaries}

\subsection{Notation}

The traditional Reinforcement Learning problem is considered, where agents interact with their environment at each timestep with an action to receive a reward and transition to a new state. The environment is stochastic and modeled by a Markov Decision Process (MDP). A trajectory consists of a state, an action taken by the policy $\pi_\theta$ parameterized by $\theta$, the reward due to the action, and the next state.

$$
\tau = (s_0, a_0, r_0, s_1, a_1, r_1, \ldots, s_{T-1}, a_{T-1}, r_{T-1}, s_T)
$$

 The policy $\pi_\theta$ maps a given state to a probability distribution over possible actions. $R(\tau)$ gives the total reward over a trajectory.

A common quantity considered in Reinforcement Learning algorithms is the Q value, or the expected return of taking an action $a_t$ at state $s_t$ while using policy $\pi_\theta$:

$$
Q^{\pi_\theta}(s_t, a_t) = \mathbb{E}[R(t)|s_t, a_t]
$$

The estimate for this quantity is given as $\hat{Q}^{\pi_\theta}(s_t, a_t)$. The value function gives the expected reward from a given state.

$$
V^{\pi_\theta}(s_t) = \mathbb{E}[R(t)|s_t]
$$

The advantage function is given by 
$$
A^{\pi_\theta}(s_t, a_t) = Q^{\pi_\theta}(s_t, a_t) - V^{\pi_\theta}(s_t)
$$

The estimated advantage function is given as $\hat{A}^{\pi_\theta}(s_t, a_t)$. The reinforcement learning problem seeks to maximize:

$$
{\eta}(\pi) = \mathbb{E}_{s_0, a_0, ...}\left[\sum_{t=0}^{\infty}{\gamma}^t r(s_t)\right]
$$

\subsection{Policy Gradient Methods}

As proven in ~\citep{Sutton:1999:PGM:3009657.3009806}, the Policy Gradient theorem allows for a convenient method through which we can compute gradients and optimize our policy.

$$ \nabla_\theta \mathbb{E}_{\tau \sim \pi_\theta}[R(\tau)] = \mathbb{E}_{\tau \sim
\pi_\theta} \left[R(\tau) \cdot \nabla_\theta \left(\sum_{t=0}^{T-1}\log
\pi_\theta(a_t|s_t)\right)\right] $$

However, this gradient needs to be computed by sampling trajectories from the environment. These trajectories have extremely high variance as each transition between states is probabilistic (the environment is formulated as a Markov Decision Process). A \emph{baseline function} is introduced into the estimate to reduce variance while still providing an unbiased signal.

$$
\nabla_\theta \mathbb{E}_{\tau \sim \pi_\theta}[R(\tau)] 
{\approx}\; \mathbb{E}_{\tau \sim \pi_\theta} \sum_{t=0}^{T-1} \nabla_\theta \log \pi_\theta(a_t|s_t) \cdot \hat{A}(s_t,a_t)
$$

For use with automatic differentiation libraries, a surrogate loss is constructed whose gradient equals the above expression.

$$
L^{PG}=\mathbb{E}_{\tau \sim \pi_\theta}[\log \pi_\theta(a_t|s_t) \cdot \hat{A}(s_t,a_t)]
$$

Though the introduction of the baseline aids in lowering variance, a key assumption in current policy gradient methods is that the distribution of states visited does not change upon changing the policy. This causes the surrogate loss functions to only be local approximations to the expected return of a policy. Due to the surrogate loss being a local approximation, state-of-the-art methods constrain the magnitude by which the parameters of the policy network are perturbed while training.

\subsubsection{Trust Region Policy Optimization}

TRPO ~\citep{DBLP:journals/corr/SchulmanLMJA15} involves a surrogate loss function with a constraint:

$$
\max_\theta{\mathbb{E}_t[{\frac{\pi_\theta(a_t|s_t)}{\pi_{\theta_old}(a_t|s_t)}}]\hat{A}_t}
$$
subject to $$\mathbb{E}_t[KL[\pi_{\theta_old}(\cdot|s_t), \pi_{\theta}(\cdot|s_t)]] \leq \delta
$$

This surrogate loss avoids overly large parameter updates that can irreversibly damage the policy's performance. However, it requires complex second order computations and is difficult to implement.

\subsubsection{Proximal Policy Optimization}

There exist two versions of PPO ~\citep{DBLP:journals/corr/SchulmanWDRK17}: one involving a penalty based on the KL divergence between the old and new policy, and another that relies on clipping the surrogate loss function when the ratio between the old and current policy grows past a set hyper-parameter. 

$$
L^{CLIP} = \hat{\mathbb{E}}[min(r_t(\theta)\hat{A}_t, clip(r_t(\theta), 1-\epsilon, 1+\epsilon)\hat{A}_t]
$$

where

$$
r_t(\theta) = \frac{\pi_\theta(a_t|s_t)}{\pi_{\theta_{old}}(a_t|s_t)}
$$

This surrogate loss avoids overly large parameter updates that can irreversibly damage the policy's performance.

\subsection{Gaussian Processes}

As explained in ~\citep{rasmussen2004gaussian}, a Gaussian process is an infinite-dimensional generalization of a Gaussian distribution. They can be used to model distributions over possible functions in a Bayesian fashion. Gaussian Processes are parameterized by a mean function \(m(x)\) and a covariance (kernel) function \(k(x,x')\). We can express a sample function from the GP as:

$$
f(\cdot) \sim GP(m(\cdot), k(\cdot, \cdot)),
$$
which means: the function f is distributed as a GP with mean function m and covariance function k.

A set of \(n\) observations, \(\mathbf{y}=\{y_1,...,y_n\}\), can be interpreted as a single sample from a multivariate Gaussian distribution that can be paired with a GP. This set of observations are derived from a function \(f(\mathbf{x})\) with some noise variance for a set of inputs, \(X\), such that:
$$
y_i = f(\mathbf{x_i}) + \epsilon,\quad \epsilon \sim \mathcal{N}(0,\sigma^2).
$$

The kernel defines the relation between the inputs; if \(x\) is distant from \(x'\), \(k(x,x') \approx 0\). The radial basis function, as expressed in ~\citep{ebden2015gaussian}:
$$
k(\mathbf{x}, \mathbf{x}') = \sigma_f^2 e^{-\frac{1}{2} (\mathbf{x}-\mathbf{x}')^T M (\mathbf{x}-\mathbf{x}')} + \sigma_n^2 \delta_{\mathbf{x},\mathbf{x}'},
$$
is a popular choice for the kernel. Here, \(\sigma_f\) is the maximum allowable covariance, \(\sigma_n\) is the noise in the underlying function, and \(\delta_{\mathbf{x},\mathbf{x}'}\) is the Kronecker delta function. If the structure of the underlying function is unknown, the mean of the partner GP is often assumed to be zero everywhere.

The GP can have predictive power as a prior for Bayesian inference. Let \(\mathbf{y}\) be the function values of the training cases \(X\), as shown above, and let \(y_*\) be the function value of the test set input \(\mathbf{x_*}\). The kernel function yields the following three matrices:

$$
K = 
\begin{bmatrix}
k(\mathbf{x_1},\mathbf{x_1}) & k(\mathbf{x_1},\mathbf{x_2}) & \cdots & k(\mathbf{x_1},\mathbf{x_n}) \\
k(\mathbf{x_2},\mathbf{x_1})& k(\mathbf{x_2},\mathbf{x_2}) & \cdots & k(\mathbf{x_2},\mathbf{x_n}) \\
\vdots  & \vdots  & \ddots & \vdots  \\
k(\mathbf{x_n},\mathbf{x_1}) & k(\mathbf{x_n},\mathbf{x_2}) & \cdots & k(\mathbf{x_n},\mathbf{x_n}) 
\end{bmatrix}
$$
$$
K_*=
\begin{bmatrix}
k(\mathbf{x_*},\mathbf{x_1}) & k(\mathbf{x_*},\mathbf{x_2}) & \cdots & k(\mathbf{x_*},\mathbf{x_n})
\end{bmatrix}
$$
$$
K_{**}=k(\mathbf{x_*},\mathbf{x_*}).
$$

Since the GP is a generalization of the multivariate Gaussian distribution,

$$
\begin{bmatrix}
\mathbf{y} \\
y_* 
\end{bmatrix}
\sim
\mathcal{N}
\begin{pmatrix}
\mathbf{0},
\begin{bmatrix}
K & K_*^T\\
K_* & K_{**} 
\end{bmatrix}
\end{pmatrix}.
$$

One of the properties of the multivariate Gaussian distribution is that conditional probabilities also follow a Gaussian distribution, so

$$
y_* | y \sim \mathcal{N}(K_*K^{-1}\mathbf{y}, K_{**}-K_*K^{-1}K_*^T).
$$

The Gaussian Process gives the following posterior prediction distribution:

$$
\mu_{y_*} = K_*K^{-1}\mathbf{y}
$$

$$
\Sigma_{y_*} = K_{**}-K_*K^{-1}K_*^T.
$$

\section{Algorithm}

\subsection{Details}

As described above, we use a Gaussian Process to directly model the expected return of a given policy. We use a Gaussian Process with a 0 mean function and the Radial Basis Function kernel. A surrogate loss function is constructed by adding a bonus $B$ to the PPO Clipped loss function, where $B$ is the expected improvement acquisition function:

$$
L^{GPPO} = L^{CLIP} + B
$$

The expected improvement acquisition function was used because of recent results showing the effectiveness of a measure of 'curiosity' in improving performance such as in ~\cite{pathakICMl17curiosity} and ~\cite{DBLP:journals/corr/PlappertHDSCCAA17}. Experimentation showed that a lengthscale parameter of $5*10^{-4}$ and a noise parameter of $10^{-2}$ led to good policies being learned.

The PPO objective provides a lower bound, local approximation to the true return for a given policy. It also disincentivizes larger updates, as it is only a local approximation and optimizing the objective beyond a range where it is accurate hurts performance. By using the Gaussian Process to explicitly learn the expected advantage of any given set of parameters, we are essentially allowing for less conservative updates while optimizing a the policy. This process also reflects recent work done in learning loss functions, in that we use a Gaussian Process to explicitly learn the function describing the returns of parameters ~\citep{DBLP:journals/corr/abs-1802-04821}.

For training the system, following ~\cite{DBLP:journals/corr/SchulmanWDRK17}, stochastic gradient descent is conducted on a modified objective that incorporates the loss from the critic as well:

$$
L_t^{CLIP+VF+GP}(\theta) = L_t^{CLIP}(\theta) + c_1 L_t^{VF}(\theta) + B_t(\theta)
$$

It is important to note that no entropy is used to facilitate exploration; exploration is conducted solely through the expected acquisition bonus that has been added. 

\subsection{Gaussian Process Update Rule}

As shown in \cite{Kakade02approximatelyoptimal} and described in \citet{DBLP:journals/corr/SchulmanLMJA15}, the expected discounted reward from policy $\pi_\theta$ can be written in terms of a different policy $\pi_{\theta_k}$ as follows:

$$
{\eta}(\pi_\theta) = {\eta}(\pi_{\theta_k}) + \sum_{s}{\rho_{\pi_\theta} \sum_{a}\pi_{\theta}(a|s) A_{\pi_{\theta_k}}(s, a)}
$$

Where 

$$
\rho_{\pi_\theta} = \rho_{\pi_\theta}(s) = P(s_0 == s) + P(s_1 == s) + P(s_2 == s) + ...
$$

Thus, it follows that

$$
{\eta}(\pi_\theta) = {\eta}(\pi_{\theta_k}) + \sum_{t=0}^{T}{\gamma^t(\sum_{s}{P_{\pi_\theta}(s_t == s) \sum_{a}\pi_{\theta}(a|s) A_{\pi_{\theta_k}}(s, a)})}
$$

$$
= {\eta}(\pi_{\theta_k}) + \sum_{t=0}^{T}{\gamma^t}\mathbb{E}_{s \sim P_{\pi_{\theta_k}}, a \sim {\pi_{\theta_k}}}\left[ \frac{P_{\pi_\theta}(s_t == s)}{P_{\pi_{\theta_k}}(s_t == s)} \frac{\pi_{\theta}(a|s)}{\pi_{\theta_k}(a|s)} A_{\pi_{\theta_k}}(s, a) \right]
$$

Assuming that the probabilities of the actions we take are independent of the probability we are in that state:

$$
= {\eta}(\pi_{\theta_k}) + \sum_{t=0}^{T}{\gamma^t}\mathbb{E}_{s \sim P_{\pi_{\theta_k}}, a \sim {\pi_{\theta_k}}}\left[ \frac{P_{\pi_\theta}(s_t == s)}{P_{\pi_{\theta_k}} (s_t == s)} \right] \mathbb{E}_{s \sim P_{\pi_{\theta_k}}, a \sim {\pi_{\theta_k}}}\left[ \frac{\pi_{\theta}(a|s)}{\pi_{\theta_k}(a|s)} A_{\pi_{\theta_k}}(s, a) \right]
$$

Note that:

$$
\mathbb{E}_{s \sim P_{\pi_{\theta_k}}, a \sim {\pi_{\theta_k}}}\left[ \frac{P_{\pi_\theta}(s_t == s)}{P_{\pi_{\theta_k}} (s_t == s)} \right] = \sum_{s}{P_{\pi_{\theta_k}} (s_t == s) \frac{P_{\pi_\theta}(s_t == s)}{P_{\pi_{\theta_k}} (s_t == s)}}
$$

$$
= \sum_{s}{P_{\pi_{\theta_k}} (s_t == s)} = 1
$$

Thus,

$$
{\eta}(\pi_\theta) = \sum_{t=0}^{T}{\gamma^t} \mathbb{E}_{s \sim P_{\pi_{\theta_k}}, a \sim {\pi_{\theta_k}}}\left[\frac{\pi_{\theta}(a|s)}{\pi_{\theta_k}(a|s)} A_{\pi_{\theta_k}}(s, a) \right] + {\eta}(\pi_{\theta_k})
$$

$$
{\eta}(\pi_\theta) = \frac{1-\gamma^{T+1}}{1-\gamma} \mathbb{E}_{s \sim P_{\pi_{\theta_k}}, a \sim {\pi_{\theta_k}}}\left[\frac{\pi_{\theta}(a|s)}{\pi_{\theta_k}(a|s)} A_{\pi_{\theta_k}}(s, a) \right] + {\eta}(\pi_{\theta_k})
$$

This result inspires an update rule on how a Gaussian Process can be trained to predict the expected discounted reward of any given policy. The Gaussian Process shall be trained to map

$$
\theta \rightarrow \frac{1-\gamma^{T+1}}{1-\gamma} \mathbb{E}_{s \sim P_{\pi_{\theta_k}}, a \sim {\pi_{\theta_k}}}\left[\frac{\pi_{\theta}(a|s)}{\pi_{\theta_k}(a|s)} A_{\pi_{\theta_k}}(s, a) \right] + {\mu}(\pi_{\theta_k})
$$

Where ${\mu}(\pi_{\theta_k})$ represents the mean prediction of ${\eta}(\pi_{\theta_k})$ by the Gaussian Process. In the algorithm, $\pi_{\theta_k}$ represents the old policy.

\subsection{Other Implementation Details}

We built our implementation of the proposed GPPO algorithm on top of the OpenAI Baselines repository ~\citep{baselines}. We used the implementations of other algorithms in the same repository to compare GPPO's performance against current methods. The policy is represented by a Multilayer Perceptron with two hidden layers of 64 units and $tanh$ nonlinearities.

\subsection{Pseudocode}

An algorithm involving $N$ actors collecting data of $T$ timesteps, and then optimizing the above objective using the Adam optimization process is described below. The Gaussian Process needs to store each set of parameters and the mean advantage they earned, but since this can become too much and slow down the system, we limit the size of the Gaussian Process' memory to $S$ points. The most recent $S$ points are used as these are most likely the points in the neighborhood of our current parameters. Our implementation used $S = 20$ and $T = 10^6$.

\begin{algorithm}
\For {iteration = 1, 2, ...}{
    \For{actor=1, 2, ... N}{
        Run $\pi_\theta$ in environment for $T$ timesteps\;
        Compute advantage estimates $\hat{A}_1, ..., \hat{A}_t$\;
        Center each $\hat{A}_i$ by subtracting the mean $\hat{A}^{m}$ from each $\hat{A}_i$\;
        Add $(\theta, \hat{A}^{m})$ to the buffer containing the Gaussian Process memory.
        
        \If{Gaussian Process memory is above size $S$}{
            remove the oldest sample from the Gaussian Process memory.\;
        }
    }
    Optimize $L^{CLIP+VF+GP}$ wrt $\theta$ for $K$ epochs and $M$ minibatches using the Adam optimizer\;
}
\caption{GPPO algorithm}
\end{algorithm}

\section{Experiments}

We compare our algorithm's performance to current state-of-the-art methods in the literature on OpenAI environments ~\citep{1606.01540}. Specifically, we use 6 included MuJoCo environments, which feature continuous action spaces and simulate robotic locomotion: Ant, HalfCheetah, Hopper, InvertedDoublePendulum, InvertedPendulum, and Swimmer, all v2. For each environment, we run the system with 3 random seeds and present results averaged across those same seeds.

\begin{figure}[h!]
    \includegraphics[width=\linewidth]{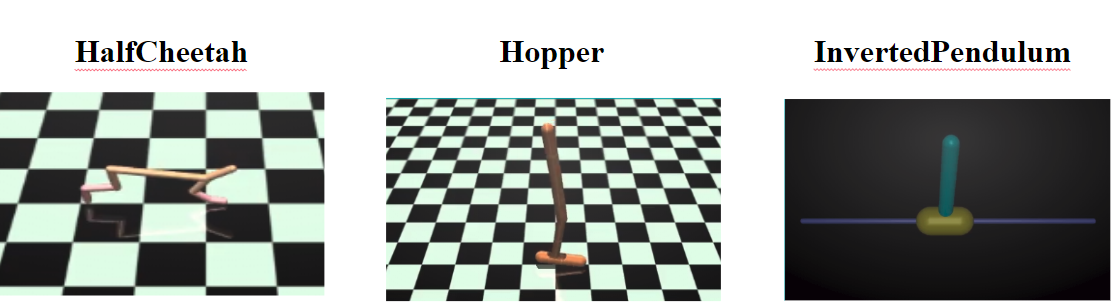}
    \caption{Some of the environments used for evaluating algorithm's performance}
    \label{fig:roboticsEnvs}
\end{figure}

\subsection{Results}

\begin{figure}[h!]
  \centering
  \subfloat[Ant-v2]{\label{ref_label1}\includegraphics[width=0.33\textwidth]{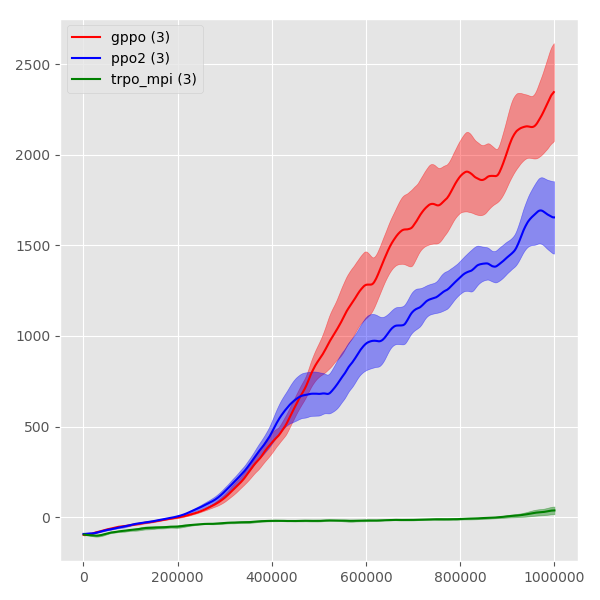}}
  \subfloat[HalfCheetah-v2]{\label{ref_label2}\includegraphics[width=0.33\textwidth]{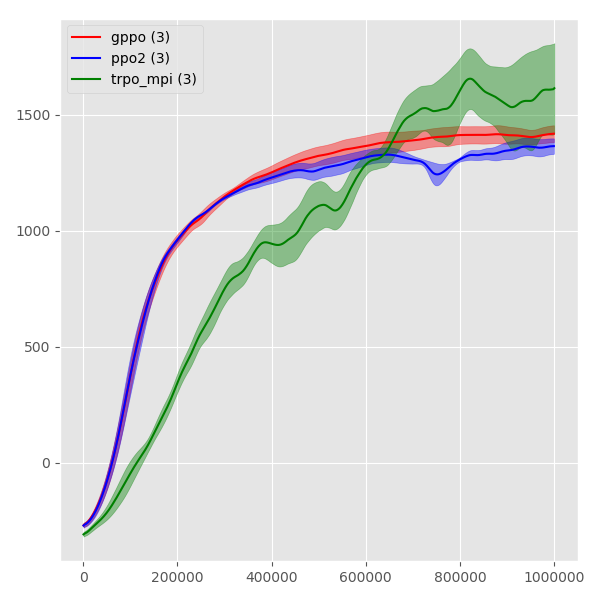}}
  \subfloat[Hopper-v2]{\label{ref_label3}\includegraphics[width=0.33\textwidth]{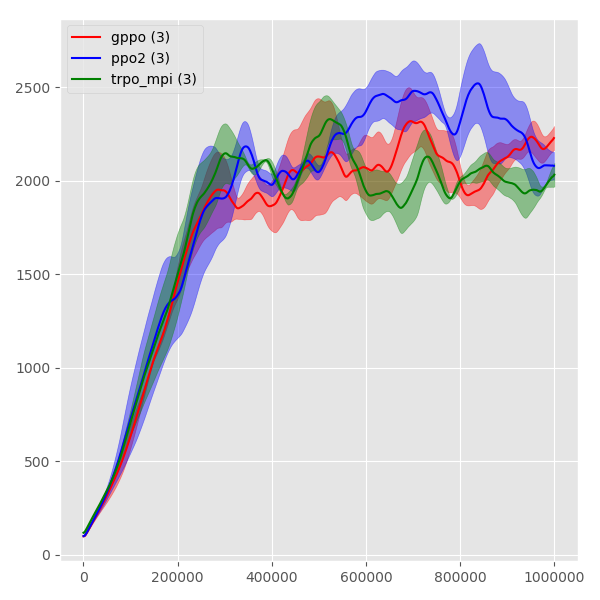}} \newline
  \subfloat[InvertedDoublePendulum-v2]{\label{ref_label4}\includegraphics[width=0.33\textwidth]{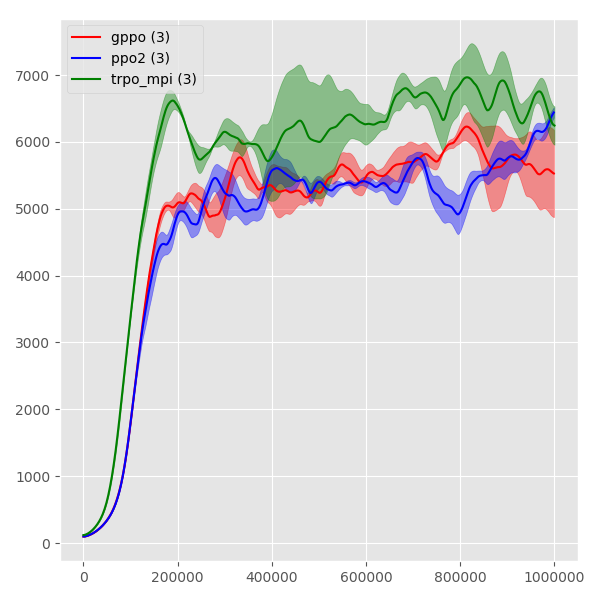}}
  \subfloat[InvertedPendulum-v2]{\label{ref_label5}\includegraphics[width=0.33\textwidth]{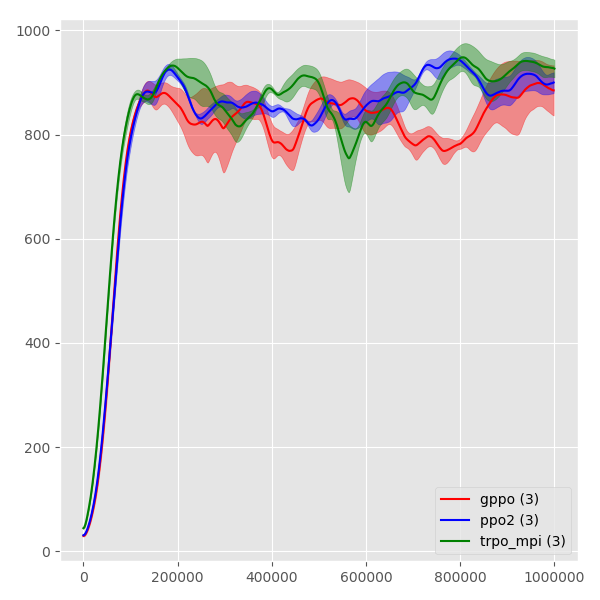}}
  \subfloat[Swimmer-v2]{\label{ref_label6}\includegraphics[width=0.33\textwidth]{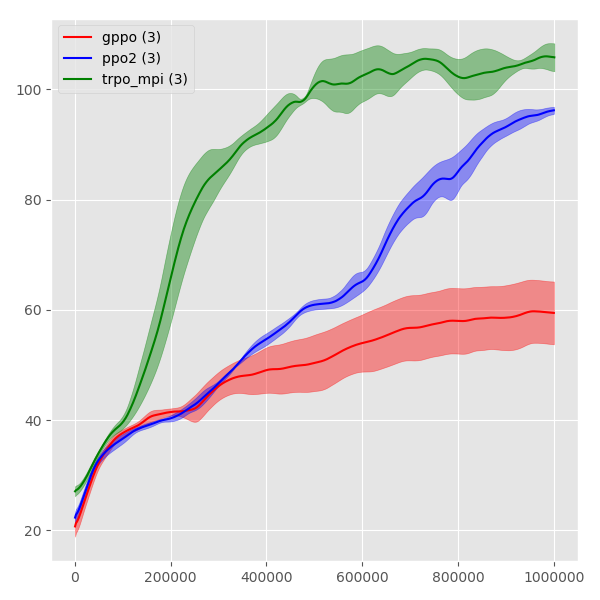}}
  \caption{\label{ref_label_overall}Episode reward over time (averaged across three trials)}
\end{figure}

The above graphs present the reward of each episode over time averaged across three trials. The table below gives the mean episode reward across the algorithm's training. A noise parameter of 1e-2 and a lengthscale of 5e-4 were used. As shown, GPPO gives comparable, and at times, superior results to current methods.

\begin{table}[h!]
\centering
\begin{tabular}{llll}
Environment               & GPPO     & PPO      & TRPO     \\
Ant-v2                    & 261.813  & 231.259  & 26.366   \\
HalfCheetah-v2            & 1119.417 & 1076.696 & 958.878  \\
Hopper-v2                 & 1237.472 & 1308.940 & 1296.983 \\
InvertedDoublePendulum-v2 & 1706.657 & 1695.973 & 2306.271 \\
InvertedPendulum-v2       & 500.694  & 525.656  & 621.293  \\
Swimmer-v2                & 49.340   & 62.348   & 86.588  
\end{tabular}
\caption{Mean return of all episodes averaged across three trials}
\end{table}

\newpage

\section{Future Work}

Several improvements could be made over the GPPO algorithm presented. For one, as described in ~\cite{hensman2013gaussian}, instead of our method of limiting the Gaussian Process 'memory' to just the recent samples, the exact inducing points that are most relevant to the predictions can be used. Additionally, performance is dependent on good parameters being set for the lengthscale and noise parameter. Some method to dynamically learn these quantities while training in the environment would likely significantly boost performance. Further experimentation with different kernels may also yield a more precise kernel that reflects the structure of a deep neural network model.

\newpage
\vskip 0.2in
\bibliography{citations.bib}

\end{document}